\documentclass[
 amsmath,amssymb,
 preprint,
 superscriptaddress
]{article}

\usepackage{amsmath}
\usepackage{amssymb}
\usepackage[english]{babel}
\usepackage{graphicx}
\usepackage{bm}
\usepackage{epsfig}
\usepackage[svgnames]{xcolor}
\usepackage{pstricks,graphics,pst-grad,times}
\usepackage{listings}
\usepackage{microtype}
\usepackage{pifont}
\usepackage{authblk}
\usepackage{blindtext}
\usepackage{hyperref}
\usepackage{threeparttable} 
\usepackage{booktabs}       
\usepackage{geometry}
\usepackage{multirow}
\title{A Note on Statistically Accurate Tabular Data Generation Using Large Language Models}

\author{Andrey Sidorenko}
\affil{\texttt{andrey.sidorenko@mostly.ai}\vspace{2ex}\\
MOSTLY AI}

\date{}

\begin{document}

\maketitle
\begin{abstract}
Large language models (LLMs) have shown promise in synthetic tabular data generation, yet existing methods struggle to preserve complex feature dependencies, particularly among categorical variables. This work introduces a probability-driven prompting approach that leverages LLMs to estimate conditional distributions, enabling more accurate and scalable data synthesis. The results highlight the potential of prompting probability distributions to enhance the statistical fidelity of LLM-generated tabular data.

\end{abstract}


\section{Introduction}
\label{sect:introduction}
Tabular data is a fundamental format in many industries, including finance, healthcare, and e-commerce. However, accessing high-quality tabular data is often challenging due to privacy constraints, data scarcity, and imbalances in real-world datasets. Synthetic data generation has emerged as a viable solution to these challenges, aiming to produce realistic datasets that preserve privacy and retain the statistical properties of the original data.

Recent advances in large language models (LLMs), particularly transformer-based architectures, have demonstrated exceptional generative capabilities in text. On the other hand, pre-trained LLMs have also shown potential in synthetic tabular data generation, using their ability to model complex distributions without requiring extensive feature engineering \cite{xu2025llmsbadsynthetictable,shi2025comprehensivesurveysynthetictabular, zhao2025surveylargelanguagemodels, fang2024large, zhao2025tabulaharnessinglanguagemodels, zhang2024aigtaigenerativetable, berkovitch2024generatingtablesparametricknowledge}. Unlike traditional synthetic data generation techniques, such as generative adversarial networks (GANs), variational autoencoders (VAEs), tabular auto-regressive networks, and copula-based models, LLMs leverage natural language processing techniques to generate structured data \cite{NEURIPS2019_254ed7d2, Park_2018, esteban2017realvaluedmedicaltimeseries, RePEc:arx:papers:1910.09504, tiwald2025tabularargnflexibleefficientautoregressive}. This text-based representation of LLMs
makes them a promising alternative for flexible and scalable tabular data generation from scratch \cite{zhao2025surveylargelanguagemodels}, especially when fine-tuning for domain-specific tabular datasets is not required.

However, despite their generative power, LLMs face several limitations in modeling tabular data. One of the primary challenges is capturing complex feature dependencies without explicit fine-tuning. While traditional methods like GANs and VAEs are designed to model joint distributions of structured data, LLMs operate auto-regressively, generating data sequentially rather than holistically resulting in not correct correlations between features. This limitation can lead to the generation of unrealistic or statistically inconsistent values, especially categorical ones, thereby compromising the utility of synthetic data for downstream tasks \cite{borisov2023language}.

To address these challenges, recent research has explored methods to improve the generation of categorical values by LLMs. For instance, the GReaT framework \cite{borisov2023language} and its advanced version GReaTER \cite{kwok2025greatergeneraterealistictabular} aim to leverage auto-regressive LLMs to sample synthetic tabular data, demonstrating improved realism in the generated datasets. Similarly, the LLM-TabFlow approach integrates LLM reasoning with score-based diffusion models to better preserve inter-column logical relationships, thereby enhancing the fidelity of categorical data generation \cite{long2025llmtabflow}.

Another promising direction involves the use of effective prompting strategies. The EPIC framework, for example, employs balanced and grouped data samples with unique variable mappings to guide LLMs in generating accurate synthetic data across all classes, even in imbalanced datasets \cite{kim2024epic}. These approaches highlight the potential of combining LLMs with tailored prompting techniques to improve the generation of categorical values in synthetic tabular data.

Despite these advancements, challenges remain. LLMs may still exhibit biases inherited from auto-regressive nature of token-by-token generation, leading to skewed representations of certain categories, rather reflecting language knowledge of LLMs than realistic distributions of categories. Additionally, the auto-regressive nature of LLMs can result in the propagation of errors, especially when generating sequences of dependent categorical variables. Addressing these issues is crucial for the development of robust and reliable tabular synthetic data generation methods.

In this study, our objective is to improve the generation of synthetic tabular data based on LLM by focusing on the accurate modeling of categorical variables. We propose a probability-driven prompting approach that estimates the distribution of categorical values and uses these estimates to generate synthetic data. This method seeks to preserve the statistical properties and interdependencies of the original data, producing more realistic and useful synthetic datasets for various applications.

\section{Background}
\label{sect:background}

Traditionally, two primary approaches have been employed to generate tabular data using LLMs (see, e.g., \cite{berkovitch2024generatingtablesparametricknowledge}). The first method, table-wide prompting, involves prompting an LLM to generate an entire table in one go, often favoring common patterns seen in training data. The model receives a request specifying the necessary columns and the desired number of records, producing a complete dataset in response. This is a fast method, but it lacks fine control over individual feature distributions and, in general, is not suitable for generating very large tables.

The second method, cell-by-cell generation, prompts an LLM to generate each cell individually, taking into account the surrounding context. Each column value is sampled sequentially, with previously generated values forming the context for subsequent features. Although this approach allows for better control over feature dependencies, it incurs significant computational overhead.

The main weakness of these approaches for tabular data generation is their inability to correctly represent real-life distributions due to the way LLMs generate text. LLMs operate in an auto-regressive manner, generating token-by-token outputs based on the probability distribution conditioned on the preceding tokens. This probability factorization follows the chain rule of probability, such that the likelihood of a sequence of tokens  \((w_1, w_2, ..., w_n)\) is computed as:
\[
p(w_1, ..., w_n) = \prod_{k=1}^{n} p(w_k | w_1, ..., w_{k-1})
\]
This approach is particularly useful for generating natural-language sequences, as it enables LLMs to generate text dynamically while maintaining contextual coherence. The model progressively refines its predictions at each step by leveraging learned patterns in the training corpus.

This fundamental mechanism introduces multiple limitations for tabular data generation. Because LLMs are optimized for language-based sequences rather than structured tabular data, they prioritize linguistic coherence over statistical accuracy, often leading to unrealistic distributions. Additionally, token bias is an inherent issue since LLMs assign probabilities to tokens based on their frequency in language data rather than the statistical distributions present in real-world tabular datasets, resulting in skewed sampling where certain categories may be over- or under-represented. Moreover, table-wide prompting lacks the ability to dynamically adjust probabilities based on inter-feature dependencies, which means that attributes may be generated independently, failing to capture real-world correlations. Finally, cell-by-cell generation exponentially increases complexity as each feature must be generated sequentially with its prior context, making large-scale dataset generation computationally infeasible. These limitations make traditional LLM-based tabular data generation methods ineffective for generating large and realistic datasets.

\section{Results and Discussion}
\label{sect:discussion}

To test these approaches, we chose a very simple real-life case of the very diverse population in California and its distribution by age and ethnicity groups. According to estimates from the US Census Bureau (for July 1, 2023) \cite{ppic2023population}, none of the ethnicity groups represents a majority of the state population (see Figure \ref{fig:uscensus} and Table \ref{table_population} in Appendix \ref{AppendixA}). In addition, the distribution of ethnicity groups is very non-uniform and depends on the age group. The original distribution shows a declining proportion of "Latino" individuals and an increasing proportion of "White" individuals with age, alongside relatively stable but modest representation of "Asian/Pacific Islander" and "Black" populations.

\begin{figure}[!h]
\includegraphics[width=12cm]{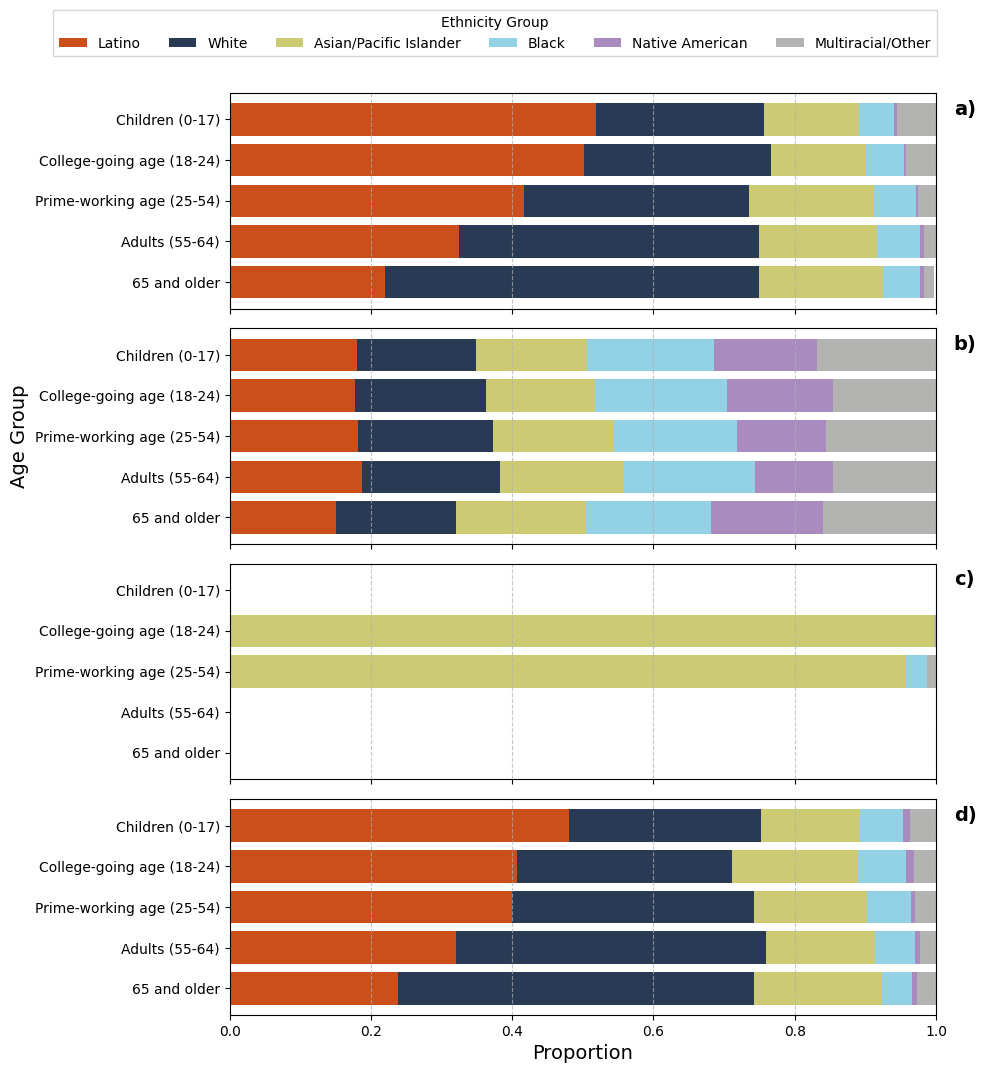}
\centering
\caption{Comparison of ethnic composition by age group in California across real and synthetic datasets. (a) Empirical distribution from the US Census Bureau (July 1, 2023) \cite{ppic2023population}; (b) Synthetic distribution using a table-wide generation approach; (c) Synthetic distribution via a cell-by-cell generation method; (d) Synthetic distribution using a probability-driven prompting approach (this work). The probability-driven prompting model preserves age-dependent demographic heterogeneity most correctly relative to the real distribution, outperforming traditional generation strategies in both accuracy and diversity of representation.}
\label{fig:uscensus}
\end{figure}

Here, we try to reproduce these statistics using different methods of prompting OpenAI's \textit{gpt-4o} model (see the prompts for the table-wide \ref{listing:table-wide-prompt} and cell-by-cell \ref{listing:cell-by-cell-prompt} generations in Appendix \ref{AppendixB}) accessed through the OpenAI API without fine-tuning.\footnote{The corresponding scripts and generated data can be found in \hyperlink{https://github.com/mostly-ai/paper-DataLLM-materials}{https://github.com/mostly-ai/paper-DataLLM-materials}} The simple table with 10,000 rows and three columns \textit{State}, \textit{Age Group}, and \textit{Ethnicity Group} was generated five times using each generation method. In this simple dataset, the \textit{State} feature is represented only by one category "California", whereas the possible values in \textit{Age Group} are "Children (0-17)", "College-going age (18–24)", "Prime-working age (25–54)", "Adults (55–64)", "65 and older", and for \textit{Ethnicity Group} the categories are "Latino", "White", "Asian/Pacific Islander", "Black", "Native American", "Multiracial/Other".

As can be seen in Fig. \ref{fig:uscensus}, the table-wide generation approach (panel b) broadly mimics the structure of the original data, but exhibits slightly increased variance and over-smoothing across age groups, especially among underrepresented minorities. In contrast, the cell-by-cell method (panel c) fails to preserve population heterogeneity and demographic granularity, generating uniform values across all age and ethnicity groups, except for a subtle presence of "Asian" and "Black" categories.

To better reproduce the behavior of the original data, we propose a probability-driven prompting approach. It involves three key steps. First, the model receives a structured prompt that defines the dataset's context, including relevant variables and categories (see Listing \ref{listing:distribution-prompt} in Appendix \ref{AppendixB}). The model then predicts probability distributions for different categories based on the given context. Finally, these probabilities are used to randomly sample values for each row in a way that maintains realistic correlations between different attributes (see the pseudo-code in Listing \ref{listing:pseudocode}). For example, in our California demographic data experiment, the new approach demonstrates its ability to capture age-dependent ethnicity distributions (see Fig.\ref{fig:uscensus}d). By first generating age groups and then conditioning ethnicity probabilities on these age groups, the system produces synthetic data that better reflects the real-world correlation between age and ethnicity in California's population, especially for dominant ("Latino" and "White") and minority groups, thus offering a superior balance between realism and diversity.

In addition to offering improved statistical accuracy, the proposed probability-driven prompting method is significantly more efficient and computationally scalable. For example, in the demographic case study based on California, where the value for the \textit{State} column is fixed to "California", the model requires only a single prompt to generate the full distribution of either age groups or ethnicity groups. Once the first feature distribution (e.g., age groups) is obtained, only five or six additional prompts are necessary to obtain conditional probabilities for the second feature (e.g., ethnicity conditioned on age group).

Crucially, this means that regardless of the number of rows being generated - whether thousands, millions, or even billions - the number of LLM invocations remains constant at five-six distributional queries. All subsequent data rows are produced via efficient sampling from the derived categorical distributions, eliminating the need for repeated LLM calls during the actual row generation process.

This efficiency stands in sharp contrast to the cell-by-cell generation approach, where each record and each cell often require a separate query, leading to substantial computational costs. While it is true that increasing the number of columns or introducing more granular categories can enlarge the distributional search space, thus potentially increasing the number of required prompts, this method still scales far more favorably. In such cases, although the number of required queries may asymptotically approach that of the cell-by-cell generation strategy, the cost remains bounded to distribution-level queries rather than record-level generation.

Consequently, this method achieves a practical balance between accuracy and computational efficiency, making it especially suitable for high-volume synthetic data generation tasks where both scalability and realism are critical. It leverages LLM capabilities for initial distribution inference while offloading the majority of computational load to lightweight post-processing via statistical sampling.

\section{Summary}
\label{sect:summary}

We demonstrated that the use of conditional probability-based generation enables a more realistic and efficient synthetic data creation process. Unlike traditional auto-regressive token-by-token generation, where each value is predicted sequentially based solely on prior tokens, this method leverages the implicit knowledge encoded within the pre-trained LLM, allowing for the estimation of coherent and statistically plausible distributions. As a result, the generated datasets better preserve natural correlations and marginal distributions, thereby enhancing their utility for downstream analytical and machine learning tasks.

Furthermore, this approach exhibits a high degree of flexibility and adaptability. It can accommodate new categories or unseen feature values while maintaining logical consistency in the data - within the bounds of the model's pre-training. If needed, the method can also be further fine-tuned on domain-specific datasets, enabling precise control over feature semantics and ensuring alignment with particular data generation objectives. Ultimately, this makes the method an interesting solution for scalable, accurate, and customizable synthetic tabular data generation.


\newpage

\bibliographystyle{unsrt}  
\bibliography{biblio}

\newpage

\onecolumn
\appendix

\section{Population distribution in California, US.}
\label{AppendixA}

\begin{table}[htbp]
\centering
\caption{Population Distribution by Age and Ethnicity Groups (\%)}
\label{table_population}
\begin{tabular}{llrrrr}
\toprule
\textbf{Age Group} & \textbf{Ethnicity Group} & \textbf{Original \cite{ppic2023population}} & \textbf{This work} & \textbf{Table-wide} & \textbf{Cell-by-cell} \\
\midrule
\multirow{6}{*}{Children (0--17)} 
& Latino & $51.9$ & $48.1 \pm 1.8$ & $18.0 \pm 0.7$ & $-$ \\
& White & $23.8$ & $27.1 \pm 1.3$ & $16.8 \pm 1.0$  & $-$ \\
& Asian/Pacific Islander & $13.4$ & $14.0 \pm 1.3$ & $15.8 \pm 0.4$  & $-$ \\
& Black & $5$ & $6.1 \pm 0.7$ & $18.0 \pm 0.8$  & $-$ \\
& Native American & $0.4$ & $1.0 \pm 0.7$ & $14.5 \pm 0.5$  & $-$ \\
& Multiracial/Other & $5.5$ & $3.7 \pm 0.9$ & $16.8 \pm 0.9$  & $-$ \\
\midrule
\multirow{6}{*}{College-going age (18--24)} 
& Latino & $50.2$ & $40.7 \pm 1.3$ & $17.7 \pm 0.9$ & $0.04$ \\
& White & $26.4$ & $30.4 \pm 2.2$ & $18.5 \pm 1.1$ & $-$ \\
& Asian/Pacific Islander & $13.5$ & $17.7 \pm 1.9$ & $15.4 \pm 0.3$ & $95.7 \pm 0.1$ \\
& Black & $5.3$ & $6.8 \pm 0.9$ & $18.8 \pm 1.0$ & $3.1 \pm 0.1$ \\
& Native American & $0.4$ & $1.1 \pm 0.7$ & $14.9 \pm 0.6$ & $-$ \\
& Multiracial/Other & $4.2$ & $3.2 \pm 0.9$ & $14.6 \pm 0.7$ & $1.2 \pm 0.2$ \\
\midrule
\multirow{6}{*}{Prime-working age (25--54)} 
& Latino & $41.7$ & $40.0 \pm 2.2$ & $18.1 \pm 0.5$ & $-$ \\
& White & $31.8$ & $34.3 \pm 3.0$ & $19.2 \pm 0.9$ & $-$ \\
& Asian/Pacific Islander & $17.7$ & $16.0 \pm 0.9$ & $17.1 \pm 0.4$ & $99.7 \pm 0.2$ \\
& Black & $5.9$ & $6.2 \pm 0.7$ & $17.5 \pm 1.1$ & $0.1 \pm 0.1$ \\
& Native American & $0.4$ & $0.6 \pm 0.3$ & $12.6 \pm 0.2$ & $-$ \\
& Multiracial/Other & $2.6$ & $3.0 \pm 1.0$ & $15.6 \pm 0.9$ & $0.2 \pm 0.1$ \\
\midrule
\multirow{6}{*}{Adults (55--64)} 
& Latino & $32.5$ & $32.0 \pm 1.5$ & $18.7 \pm 1.4$  & $-$ \\
& White & $42.4$ & $43.9 \pm 2.3$ & $19.5 \pm 0.4$  & $-$ \\
& Asian/Pacific Islander & $16.7$ & $15.4 \pm 1.8$ & $17.6 \pm 0.4$  & $-$ \\
& Black & $6.2$ & $5.7 \pm 0.3$ & $18.6 \pm 1.3$  & $-$ \\
& Native American & $0.5$ & $0.7 \pm 0.5$ & $11.0 \pm 0.5$  & $-$ \\
& Multiracial/Other & $1.7$ & $2.3 \pm 0.4$ & $14.5 \pm 0.8$  & $-$ \\
\midrule
\multirow{6}{*}{65 and older} 
& Latino & $22$ & $23.9 \pm 3.4$ & $15.1 \pm 1.1$  & $-$ \\
& White & $53$ & $50.4 \pm 5.8$ & $17.0 \pm 0.7$  & $-$ \\
& Asian/Pacific Islander & $17.5$ & $18.1 \pm 3.7$ & $18.4 \pm 0.8$  & $-$ \\
& Black & $5.3$ & $4.3 \pm 1.0$ & $17.7 \pm 1.3$  & $-$ \\
& Native American & $0.5$ & $0.7 \pm 0.2$ & $15.9 \pm 1.1$  & $-$ \\
& Multiracial/Other & $1.4$ & $2.6 \pm 0.3$ & $15.9 \pm 0.6$  & $-$ \\
\bottomrule
\end{tabular}
\end{table}

\section{Prompts}
\label{AppendixB}
\begin{lstlisting}[caption={Table-wide generation prompt},label={listing:table-wide-prompt},captionpos=b]
'''
Generate a table with exactly {SAMPLE_SIZE} records with columns
'State', 'Age Group', and 'Ethnicity Group'.

'State' contains identical values, all set to 'California/CA'.

'Age Group' should be sampled from the categories 
'{target_features['Age Group']['categories']}'.

'Ethnicity Group' should be sampled from the categories 
'{target_features['Ethnicity Group']['categories']}' reflecting 
population in 'State' of California/CA.

Only return the JSON object. Do not include any additional text, 
explanations, or formatting.\n
'''
\end{lstlisting}

\begin{lstlisting}[caption={Cell-by-cell generation prompt template},label={listing:cell-by-cell-prompt},captionpos=b]
'''
Based on the provided context and data description, generate one 
random sample for the column {user_prompt}.

Sample from the following categories: {categories}
# Context: {context}
# Data Description: {description}

The output should be limited strictly to the chosen category 
without any additional explanations or formatting.

# Response:
'''
\end{lstlisting}

\begin{lstlisting}[caption={Probability-based distribution generation prompt template},label={listing:distribution-prompt},captionpos=b]
'''
Based on the provided context and data description, generate one random sample
for the column {user_prompt}.
Sample from the provided categories

# Categories: {categories}
# Context: {context}
# Data Description: {description}

The output should be limited strictly to the JSON structure without any 
additional explanations or formatting.
'''
\end{lstlisting}

\section{Pseudo-code for probability-based distribution generation}
\label{AppendixC}
The pseudo-code of the main function that orchestrates the data generation process for all specified target features is shown in Listing \ref{listing:pseudocode}. The code creates probabilistic distributions for different demographic features based on context provided, and then samples from these distributions to create coherent datasets.
Workflow:
\begin{itemize}
    \item Process each feature defined in the ${target\_categories}$
    \item Get probability distributions or ranges from LLMs
    \item Generate samples based on these distributions
    \item Build contextual information as features are generated
\end{itemize}
\begin{lstlisting}[caption={Pseudocode for probabilistic generation of categoical tabular data},label={listing:pseudocode},captionpos=b]
Initialize generated_results
Create contexts from seed data or empty strings

FOR EACH feature_name, feature_prop IN target_categories:
    Initialize samples list
    Initialize LLM outputs storage    
    Use provided categories
    
    Create unique hashes for each context
    
    IF multiple categories exist:
        Create prompts for probability estimation
        Try to get LLM outputs with retries
        Validate and normalize probabilities
        Sample from distributions for each context
        Convert numeric ranges to actual values if needed
    ELSE IF only one category exists:
        Repeat that category for all samples
    
    Update contexts with new feature values
    Store generated values in results
\end{lstlisting}

\end{document}